\documentclass{article}

\usepackage[preprint, nonatbib]{neurips_2022}

\usepackage[utf8]{inputenc} 
\usepackage[T1]{fontenc}    
\usepackage[hidelinks]{hyperref}       
\usepackage{url}            
\usepackage{booktabs}       
\usepackage{amsfonts}       
\usepackage{nicefrac}       
\usepackage{microtype}      
\usepackage{xcolor}         

\usepackage{graphicx}
\usepackage{amsmath}
\usepackage{xspace}

\newcommand{\env}[1]{\textit{#1}}

\newcommand{\rsq}[0]{R$^2$\xspace}

\title{Atari-5: Distilling the Arcade Learning Environment down to Five Games}

\author{%
  Matthew Aitchison\\
  Australian National University\\
  \texttt{matthew.aitchison@anu.edu.au} \\
  \And
  Penny Sweetser\\
  Australian National University\\
  \\
  \And
  Marcus Hutter\\
  Australian National University / Deepmind
}

\begin{document}

\maketitle

\begin{abstract}
The Arcade Learning Environment (ALE) has become an essential benchmark for assessing the performance of reinforcement learning algorithms. However, the computational cost of generating results on the entire 57-game dataset limits ALE's use and makes the reproducibility of many results infeasible. We propose a novel solution to this problem in the form of a principled methodology for selecting small but representative subsets of environments within a benchmark suite. We applied our method to identify a subset of five ALE games, called \mbox{\textit{Atari-5}}, which produces 57-game median score estimates within 10\% of their true values. Extending the subset to 10-games recovers 80\% of the variance for log-scores for \textit{all} games within the 57-game set. We show this level of compression is possible due to a high degree of correlation between many of the games in ALE.
\end{abstract}

\section{Introduction}

The Arcade Learning Environment (ALE) \cite{bellemare13arcade} has become the gold standard for evaluating the performance of reinforcement learning (RL) algorithms on complex discrete control tasks. Since its release in 2013, the benchmark has gained thousands of citations and almost all state-of-the-art RL algorithms have featured it in their work \cite{schrittwieser2020mastering, kapturowski2018recurrent, horgan2018distributed, hessel2018rainbow, gruslys2018the, mnih2013playing}. However, results generated from the full benchmark have typically been limited to a few large research groups.\footnote{Of the 17 algorithms listed on \url{paperswithcode.com} with full Atari-57 results, only one was from a research group outside Google or DeepMind.} We posit that the cost of producing evaluations on the full dataset is not feasible for many researchers. Additionally, the verification of superiority claims through statistical tests on multiple seeds remains exceedingly rare, likely, due to the high cost.

While technological improvements have reduced computation costs, this decrease has been greatly outpaced by an increase in the scale at which RL algorithms are run. For example, the number of frames used for training to produce state-of-the-art results on ALE increased by more than \textit{one-thousand times} in the five years between 2015's Deep Q-learning model \cite{mnih2015human} and the more recent Agent57 \cite{badia2020agent57}. Training modern machine learning models can also produce non-trivial amounts of carbon emissions \cite{strubell2019energy}, for which we provide some analysis in Appendix \ref{app:cost_estimate}. Previous works have dealt with this challenge by self-selecting ad hoc subsets of games. However, this approach adds bias and makes comparison between works more challenging.

In this paper,\footnote{Source code for this paper can be found at \url{https://github.com/maitchison/Atari-5}.} we outline a principled approach to selecting subsets of environments from a RL benchmark suite. We show that, when carefully chosen, a surprisingly small subset of ALE can capture most of the useful information of a full run. We present a new benchmark, \textit{Atari-5}, which produces scores that correlate very closely to median score estimates on the full dataset, but at less than one-tenth the cost. We hope that this new benchmark will allow more researchers to participate in this important field of research, speed up the development of novel algorithms through faster iteration, and make the replication of result in RL more feasible. Our primary contributions are as follows. First, a methodology for selecting representative subsets of multi-environment RL benchmarks according to a target summary score. Second, the introduction of the Atari-5 benchmark. Finally, evidence demonstrating the high degree of correlation between scores for many games within ALE.

\begin{figure}[t]
    \centering
    \includegraphics[width=0.9\textwidth]{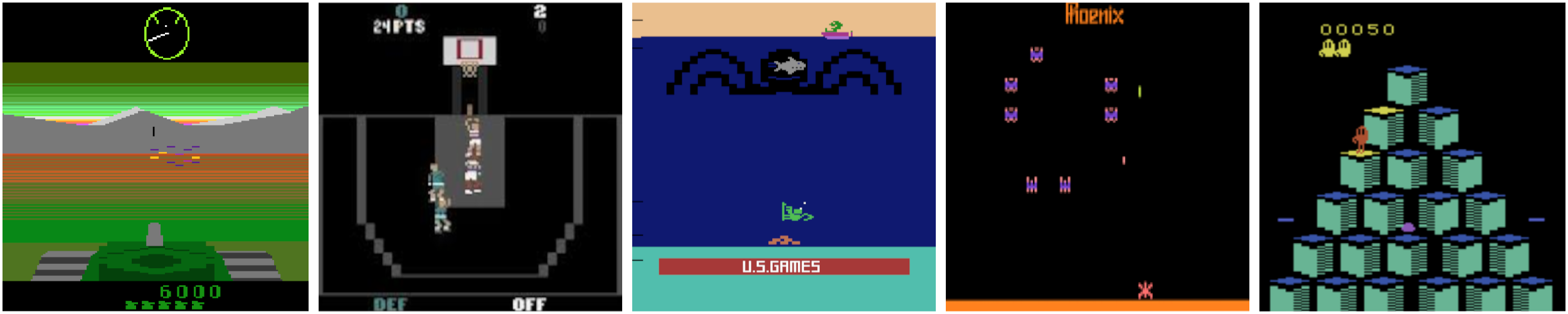}
    \caption{The five games used in the Atari-5 Subset. From left to right, \env{Battle Zone}, \env{Double Dunk}, \env{Name this Game}, \env{Phoenix}, and \env{Q*bert}.}
    \label{fig:ale}
\end{figure}

\section{Background and Related Work}

\textbf{The Arcade Learning Environment (ALE).} Despite deceptively simple graphics, ALE \cite{bellemare13arcade} provides a challenging set of environments for RL algorithms. Much of this challenge stems from the fact that ALE contains games spanning multiple genres, including sport, shooter, maze and action games, unlike most other RL benchmarks. This is important because being able to achieve goals in a wide range of environments has often been suggested as a useful definition of machine intelligence \cite{legg2007universal}. In addition, unlike many other RL benchmarks, the environments within ALE were designed explicitly for \textit{human play}, rather than \textit{machine play}. As such, ALE includes games that are extremely challenging for machines but for which we know a learnable (human) solution exists. Human reference scores also provide a means by which `good' scores can be quantified.

\textbf{ALE Subsets.} Many researchers make use of ALE subsets when presenting results. However, decisions about which games to include have varied from paper to paper. The Deep Q-learning Network (DQN) paper \cite{mnih2013playing} used a seven-game subset, while the Asynchronous Advantage Actor-Critic (A3C) paper \cite{mnih2016asynchronous} used only five of these games. A useful taxonomy was introduced by Bellemare et al. \cite{bellemare2016unifying}, which includes the often-cited \textit{hard exploration} subset. A more comprehensive list of papers and the subsets they used is given in Appendix \ref{app:subsets}, highlighting that there is currently no standard ALE subset. The key difference in our work is that our selection of games has been chosen by a principled approach to be representative of the dataset as a whole. To our knowledge, this is the first work that investigates the selection and weighting of an RL benchmark suite.

\textbf{Computationally Feasible Research.} As has been pointed out, the computational requirements for generating results on ALE can be excessive \cite{revisiting2020}. There have been several attempts to reduce this cost, including optimizations to the simulator codebase\footnote{https://github.com/mgbellemare/Arcade-Learning-Environment/pull/265}, and a GPU implementation of the environments \cite{dalton2020accelerating}. However, as these changes reduce the cost of simulating the environment and not the much higher cost of training a policy, they result in only minor improvements. Asynchronous parallel training \cite{horgan2018distributed} partially addresses this by making better use of parallel hardware but still requires access to large computer clusters to be effective. While these improvements have been helpful, they do not go far enough to make ALE tractable for many researchers. Furthermore, our subsetting approach is complementary to the approaches above.

\begin{table}[h]
    \centering
    \begin{tabular}{l l}
        \toprule
         Benchmark & Citations \\
        \midrule
         ALE \cite{bellemare13arcade} & 2,295 \\
         Vizdoom \cite{kempka2016vizdoom} & 610 \\
         DeepMind Lab \cite{beattie2016deepmind} & 380 \\
         Procgen \cite{cobbe2020leveraging} & 166 \\
         Gym Retro \cite{nichol2018retro} & 119 \\
        \bottomrule
    \end{tabular}
    \caption{Popular discrete-action vision-based benchmarks in RL by citation. Citations are as of Mar 2022.}
    \label{tab:benchmark_citations}
\end{table}


\textbf{Alternatives to ALE.} Since ALE's introduction, many other RL benchmarks have been put forward, such as \cite{cobbe2020leveraging, kempka2016vizdoom, beattie2016deepmind, nichol2018retro}. However, ALE still dominates research, and at the time of publishing, ALE has more than double the citations of these other benchmarks combined (see Table \ref{tab:benchmark_citations}). Minatar \cite{young2019minatar} addresses some of ALE's issues by creating a new benchmark inspired by Atari that presents agents with objects rather than pixels. This simplification speeds up training but requires previous algorithms to be trained on the new environments and lacks the vision part of the task, an important part of ALE's challenge.

\section{Summary Scores on Subsets of Benchmark Suites}


When working with a benchmark made up of multiple environments it is often useful to distil the performance of an algorithm, over all environments, down to a single `summary score' \cite{bellemare13arcade}. If this summary score uses scores from only a subset of the environments then the time taken to generate a summary score can be reduced. Ideally this subset summary score would:

\begin{enumerate}
    \item Preserve order.
    \item Provide useful information about the performance of the algorithm.
    \item Minimize the number of environments needed to be evaluated.
\end{enumerate}

We formalize these desiderata as follows. Let $M=\{\mu_i|i \in [1..n]\}$ be an ordered set of $n$ unique environments that make up the benchmark. For some algorithm $a$ we write the empirical score on the $i$-th environment as $a_i$. Then, for this set of environments, we say that for a pair of algorithms $a,b$, we have that $a \geq b$ if and only if $a_i \geq b_i \, \forall \, i \in [1..n]$ and that $a > b$ if there also exists some $i$, such that $a_i > b_i$. We also define a summary score $\Phi_M$ on scores from benchmark $M$ as a function $\Phi_M: \mathbb{R}^n \rightarrow \mathbb{R}$ reducing a score vector $\langle a_1, a_2, ..., a_n \rangle$ taken from benchmark $M$, to a scalar indicating performance. For convenience, we write the summary of individual scores for an algorithm, $\Phi_M(\langle a_1, a_2, ..., a_n \rangle)$ as $\Phi_M(a)$. Finally we define a subset score as a summary score which uses a strict subset of $M$.

\subsection{Order Preservation}

Ideally a subset summary score would be strictly order preserving, in that if $a > b$ then  $\Phi(a) > \Phi(b)$. Unfortunately this is not possible.

\textbf{Proposition 1:} No subset score can be strictly increasing.

\textit{Proof.} Let $M'$ be the strict subset of environments used by $\Phi_M$, then select $z \in [1..n]$ where $\mu_z \in M \setminus M'$. Now consider two algorithms $a, b$, where $a_i = b_i \; \forall \, i \neq z$, and $a_z > b_z$. Hence we have that $a > b$, but not $\Phi(a) > \Phi(b)$. Therefore, the next best thing we can do is to require that our subset score be at least \textit{weakly} order preserving, that is, if $a > b$ we can not have $\Phi_M(a) < \Phi_M(b)$.

\textbf{Proposition 2:} Any summary score of the form 

$$\Phi_M(a) = \sum_{i = 1}^{n} c_i a_i$$ 
with $c_i \in \mathbb{R} \geq 0$, is weakly increasing.


\textit{Proof.} See Appendix \ref{app:monotonic}.

We therefore find that linear combination of scores, if constrained to non-negative weights, is weakly order preserving. We note that this is not the case for more sophisticated non-linear models, such as deep neural networks, which might reverse the order of algorithms. We also note that any monotonically increasing function $\phi: \mathbb{R} \rightarrow \mathbb{R}$ can be applied to the individual game scores, which we will take advantage of by using a log transform. We can therefore guarantee that if algorithm $a$ is strictly better than algorithm $b$, it can be no worse when measured using a weighted subset score, which is, as shown, the best we can do.

\subsection{Provide Useful Information}


As shown, a weighted subset produces a monotonic measure of an algorithm's performance. However, the question of which subset to use and how to weigh the scores remains. A trivial solution would be to weigh each score by $0$, which would be fast to generate and weakly order-preserving but completely uninformative. Therefore, we propose that individual scores should be weighted such that the summary score \textit{provides meaningful information about the performance of the algorithm}. For established benchmarks, meaningful summary performance measures have often already been established. For example, with ALE, the most common summary score is \textit{median score} performance. Therefore, we suggest that the prediction of an established summary score (target score) provides a sensible strategy for selecting and weighting the environments within the suite. That is, to find the subset of environments that best predicts the target score and weigh them according to their linear regression coefficients. 

\subsection{Procedure}

We therefore propose that using linear regression models to predicate a meaningful target summary score provides a good solution to the desiderta outlined above, and formalize the procedure. %
Given $m$ algorithms $a^k:k=1...m$ together with their individual evaluations 
$s^k_i \in\mathbb{R}$ on environments $i\in\{1:n\}$ and total/aggregate score $t^k\in\mathbb{R}$,
e.g.\ $t^k=s_1^k+...+s_d^k$ or the average or the median.
The ``training'' data set is hence $D=\{(s_1^k,...,s_n^k;t^k):k=1...m\}$.
Let $I\subseteq\{1:n\}$ be a subset of games and $s^I:=(s^i:i\in I)$ be a corresponding score sub-vector.
We want to find a mapping $f^*_I:\mathbb{R}^I\to\mathbb{R}$ that best predicts $t$ from $s^I$,
i.e.\ formally $f^*_I=\arg\min_f\sum_{k=1}^m(t^k-f(s_I^k))^2$.
Since for $I=\{1:d\}$, the best $f$ is perfect and linear,
it is natural to consider linear $f$, which leads to an efficiently solvable linear regression problem.
We further want to find small $I$ with small error.
We hence further minimize over all $I\subseteq\{1:n\}$ of some fixed size $|I|=C$.
The optimal set of games $I^*$ with best linear evaluation function $f^*$ based on games in $I^*$ hence are
\begin{align*}
  (I^*,f^*) ~:=~ \mathop{\arg\min}_{|I|=C}~\mathop{\arg\min}_{f\in\text{Linear}_I} \sum_{k=1}^m(t^k-f(s_I^k))^2.
\end{align*}


\section{Application to ALE}
\label{sec:method}

This section provides details for the application of our method to the popular ALE benchmark, but note that this method could also be applied to other benchmark suites such as the Procgen \cite{cobbe2020leveraging}, MuJoCo \cite{todorov2012mujoco}, or even benchmarks outside of RL.

\subsection{Data and Processing}
\label{sec:dataset}

We used the website paperswithcode\footnote{\url{https://paperswithcode.com/dataset/arcade-learning-environment}} as the primary source of data for our experiments. This website contains scores for algorithms with published results on various benchmarks, including ALE. The dataset was then supplemented with additional results from papers not included on the website. Additions included various algorithms, such as evolutionary algorithms and shallow neural networks. A full list of algorithms and their sources contained within the dataset can be found in Appendix \ref{app:algo}.

We removed any algorithms that had results for fewer than 40 of the 57 games in the standard ALE dataset on the grounds that a reasonable median estimate could not be obtained. This reduced our dataset of 116 algorithms down to 62. When fitting models for subsets, any algorithm missing a score for a required game was excluded for that subset. We also excluded any games with results for fewer than 40 of the remaining 62 algorithms. The only game meeting this criteria was \env{Surround}. All scores were normalized using the standard approach from \cite{bellemare13arcade}
\begin{align}
    Z_i(x) := 100 \times \frac{x_i-r_i}{h_i-r_i}
\end{align} where $r_i$ and $h_i$ are the human and random scores for environment $i$ respectively, which we include in Appendix \ref{app:norm_constants}. We found that, even after normalization, scores differed across games by many orders of magnitude and produced non-normal residuals. We, therefore, applied the following transform

\begin{align}
    \phi(x)  &:= \log_{10}(1+\max(0, x))\\
    \phi^{-1}(x) &= 10^x - 1
\end{align}

to the normalized scores, producing log normalised scores.  %
%
%
Clipping was required as algorithms occasionally produce scores slightly worse than random, generating a negative normalized score. Both input and target variables were transformed using $\phi$, and all loss measures were calculated in transformed space. This transformation also has the effect that larger scoring games would not dominate the loss.

\subsection{Predictive Ability of Individual Games}

To better understand the games within ALE, we produced single-game linear models where the score from a single environment was used to predict the median Atari-57 score. We then ranked games according to their predictive ability as measured by their \rsq. Due to a small parameter count (n=2), cross-validation was not used.

\subsection{Subset Search}
\label{sec:subset_search}

To assess the best selection of games for each subset, we evaluated all subsets of the ALE games of size five three or one that met our criteria as defined in Section \ref{sec:dataset}.\footnote{We found diminishing returns after five games, however, we do also provide models for a larger ten-game set if additional precision is required.} Subsets were evaluated by fitting linear regression models to the data using the log normalized scores of games within the subset to predict the median overall, selecting the model with the lowest 10-fold cross-validated mean-squared-error. For these models, we disabled intercepts as we wanted a random policy to produce a score of 0.

To make the best use of the 57 games available, we carefully ordered the subset searches to maximize the performance of the most common subset size while maintaining the property that smaller subsets are true subsets of their larger counterparts. All subsets of five games were evaluated, with the best subset selected as Atari-5. The Atari-3 subset was then selected as the best triple game subset of the Atari-5 subset and Atari-1 as the best single-game subset of Atari-3. Doing this ensures that only five games are included in the test set and also that evaluations on Atari-1 or Atari-3 can more easily be upgraded to Atari-5 if required. For the validation subsets, we selected the best three games, excluding the games used in Atari-5, as the validation set. We then extended this set to 5 games by choosing the best two additional games. Finally, the ten-game Atari-10 set was generated by selecting the best five additional games, not in Atari-5 nor Atari-5-Val. 

Searching over all subsets took approximately 1-hour on a 12-core machine. No GPU resources were used. Source code to reproduce the results are provided in the supplementary material.

\subsection{Prediction over All Game Scores}

To better understand how much information our subsets capture about the 57 game dataset, we trained 57 linear regression models for the best five-game and ten-game subsets, predicting log-normalized scores for each game within the 57-game set. We then evaluated each of the 57 models according to their \rsq, allowing us to identify well or poorly captured environments by the subset. These models also enable score predictions for any game within the 57-game set.

\section{Results}

In this section, we present the results of our subset search on the ALE dataset using the Atari-57 median score as the target metric.

\subsection{Single Environment Performance}

\begin{figure*}[b]
    \centering
    \includegraphics[width=0.95\textwidth]{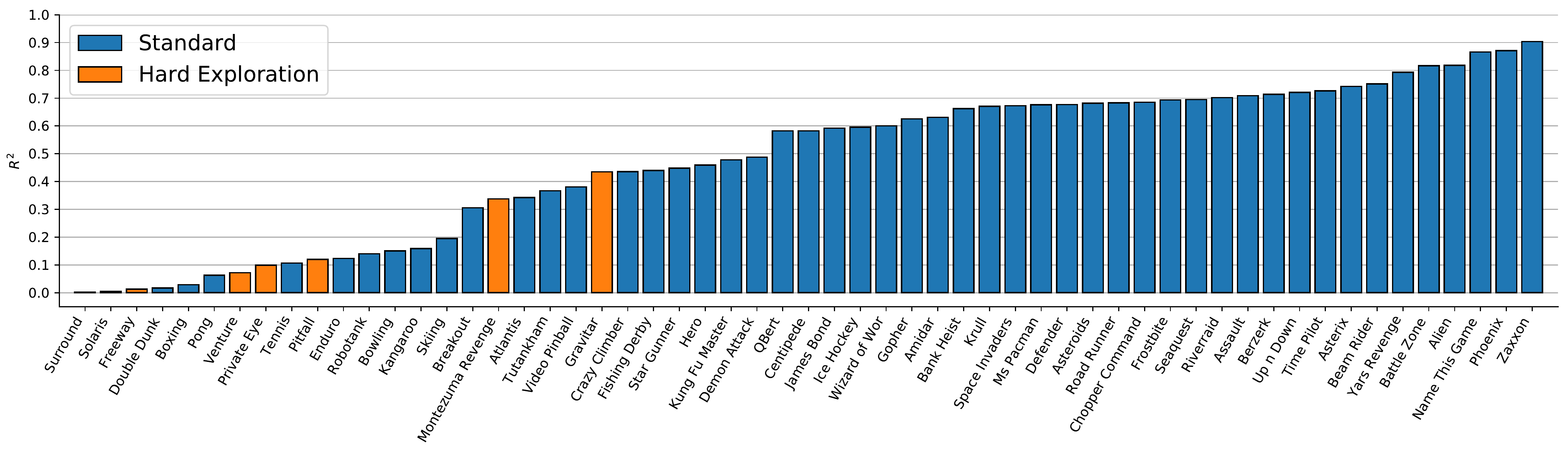}
    \caption{Not all games are good predictors of median score in the ALE. Games in the ALE benchmark sorted from least predictive (left) to most predictive (right), when used as the sole predictor for Atari-57 median performance.}
    \label{fig:single_env}
\end{figure*}

We found a significant difference in the ability of games within the ALE benchmark suite to predict median score estimates (Figure \ref{fig:single_env}).
For single-game subsets, \env{Zaxxon} performed best with $\text{R}^2=0.903$, and \env{Surround} the worst with $\text{R}^2=0.002$. The environment \env{Name this Game} was found to have the interesting property that its intercept was close to 0 (-0.03) and its scale close to 1 (1.01). As a result, normalized scores from \env{Name This Game} provide relatively good estimates for the median performance of an algorithm, by themselves, without modification.

We also note that several games have close to no predictive power when used by themselves. In the case of \textit{Pong}, \textit{Surround}, and \textit{Tennis}, this is most likely due to a low score cap, which most algorithms easily achieve. Games within the hard exploration subset also performed poorly at predicting the median score due to many algorithms achieving no better than random performance.

\subsection{Optimal Subsets}

As a baseline reference, we include results for the 7-game set used in the DQN paper \cite{mnih2015human}. The result demonstrates the importance of game selection in that our single-game model outperforms the 7-game baseline set used by the DQN paper. Table \ref{tab:performance} gives the results for the top-performing regression model for each subset size. Coefficients are provided in Appendix \ref{app:coefficents}.

We also approximated the expected relative error of the median score, using $\log_e(10)$ times the mean absolute error in log space, which, for small residuals, is a good approximation of the relative error.\footnote{The authors were unable to find a reference for this, and so have provided a proof in Appendix \ref{app:residual}.} We found that Atari-5 is generally within 10\% of the true median score. In Table \ref{tab:popular}, we give the performance of popular algorithms on the Atari-5 benchmark. 

\begin{table*}[t]
    \centering
    \resizebox{\textwidth}{!}{%
    \begin{tabular}{l l l r}
        \toprule
        Name      & Games & \rsq & $\approx$ Rel. Err. \\
        \midrule
        Atari-1 & Name This Game & 0.864 & 27.4\% \\
        Atari-3   & Battle Zone, Name This Game, Phoenix & 0.976 & 13.7\% \\
        Atari-5   & Battle Zone, Double Dunk, Name This Game, Phoenix,
        Q*Bert & 0.984 & 10.4\% \\
        Atari-10 & Amidar, Bowling, Frostbite, Kung Fu Master,  River Raid, & 0.992 & 7.2\% \\ & Battle Zone, Double Dunk, Name This Game, Phoenix, Q*Bert &  \\
        \midrule
        Atari-3-Val & Assault, Ms. Pacman, Yar's Revenge & 0.952 & 17.1\% \\
        Atari-5-Val & Bank Heist, Video Pinball, Assault, Ms. Pacman, Yar's Revenge & 0.972 & 14.3\% \\
        \midrule
        DQN-7 & Beam Rider, Breakout, Enduro, Pong, Q*Bert, Seaquest, S. Invaders & 0.756 & 38.9\%
        \\
        
        \bottomrule
    \end{tabular}
    }
    \caption{Performance, and included games, for each of the best models for each subset size.}
    \label{tab:performance}
\end{table*}

\begin{table}[h]
    \centering
    \begin{tabular}{l r r r}
        \hline
        Algorithm & Median & Atari-5 & Rel. Error \\
        \midrule
        MuZero  & 2,041 & 2,091 & 2.5\% \\
        Agent57  & 1,975 & 1,817 & 8.0\% \\
        Ape-X  & 434 & 475 & 9.6\% \\
        IQN  & 237 & 215 & 9.2\% \\
        Rainbow DQN & 227 & 225 & 0.9\% \\
        \bottomrule
    \end{tabular}
    \caption{Results for
        MuZero \cite{schrittwieser2020mastering},
        Agent57 \cite{badia2020agent57}, 
        Ape-X \cite{horgan2018distributed},
        IQN \cite{dabney2018implicit}, 
        and Rainbow DQN \cite{hessel2018rainbow}. 
        Atari-5 predictions fall close the true Atari-57 median scores, and the ranking of the algorithms is generally preserved.
    }
    \label{tab:popular}
\end{table}

\subsection{Robustness to Environmental Settings}

To assess the impact of significantly different environmental settings, we used Atari-5 to evaluate the algorithm Go-Explore \cite{ecoffet2021first}, which was not included in our training set. Go-Explore uses 400,000 frame time limits for each game, rather than the 108,000 used by the algorithms in our dataset. For many games, this difference fundamentally changes the skills the game requires and, importantly, the maximum possible score. For example, in the Space Invaders-like game \textit{Demon Attack}, a shorter time limit requires an agent to take risks to score points more quickly. In contrast, an extended time limit emphasises staying alive instead. Given the differences, we were surprised to find that Atari-5 produced an overestimate of only 12.2\% (an estimate of 1,620 compared to 1,446). While this result supports the robustness of Atari-5 to provide useful performance information, even when environmental settings have been changed, we still recommend that it not be used in these cases.

\subsection{Fairness of Atari-5 Scores}

To assess the fairness of Atari-5, we split the algorithms with scores for all five games in the Atari-5 dataset into tertiles, ordered by their true median score performance. We looked for differences in accuracy between the groups by comparing the average absolute relative error and looked for bias by comparing the average relative error. No statistically significant difference was found between any pairs using the standard p=0.05 threshold and Wilson's two-sided T-test.

\begin{figure}[h]
    \centering
    \includegraphics[width=0.23\textwidth]{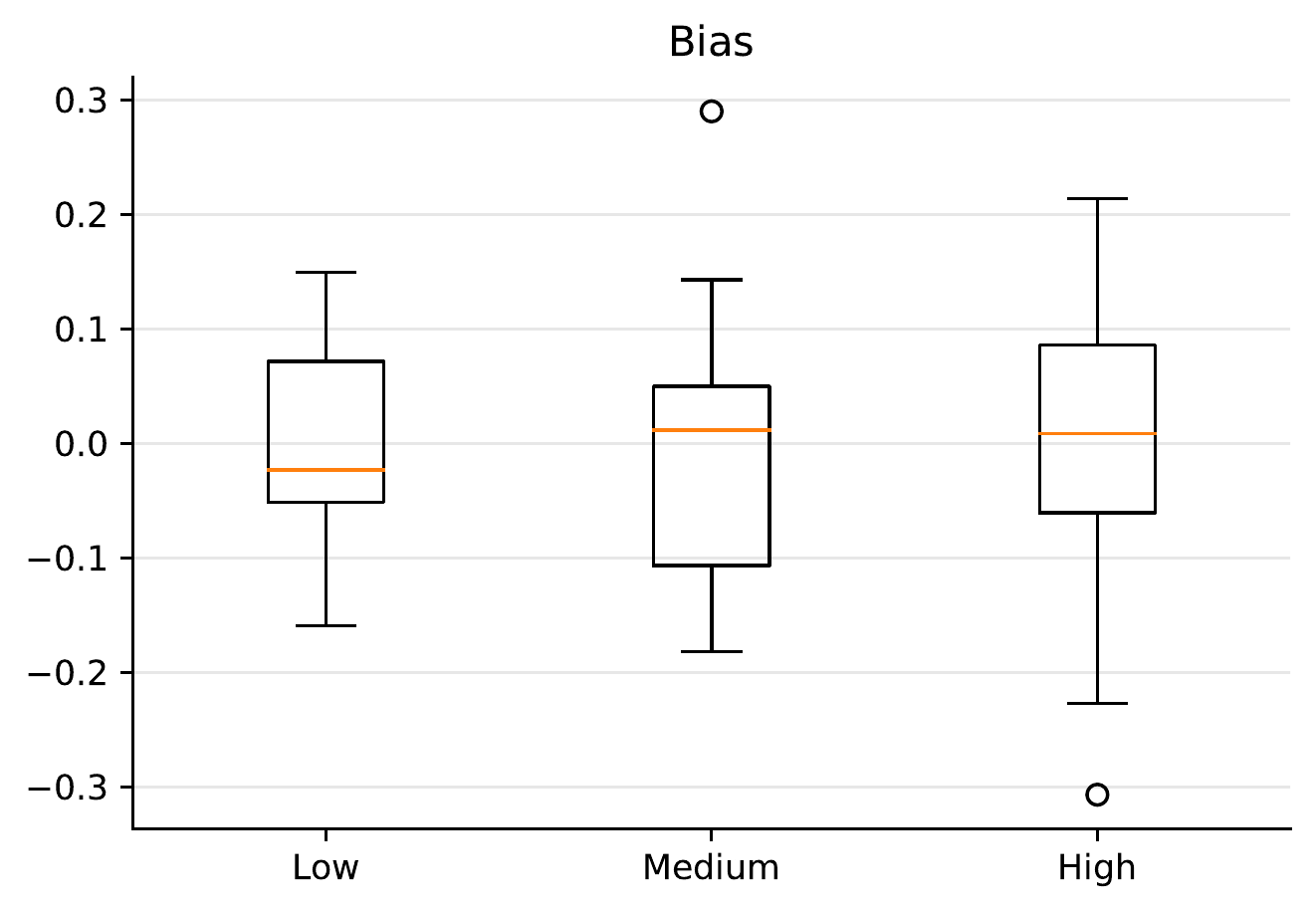}
    \includegraphics[width=0.23\textwidth]{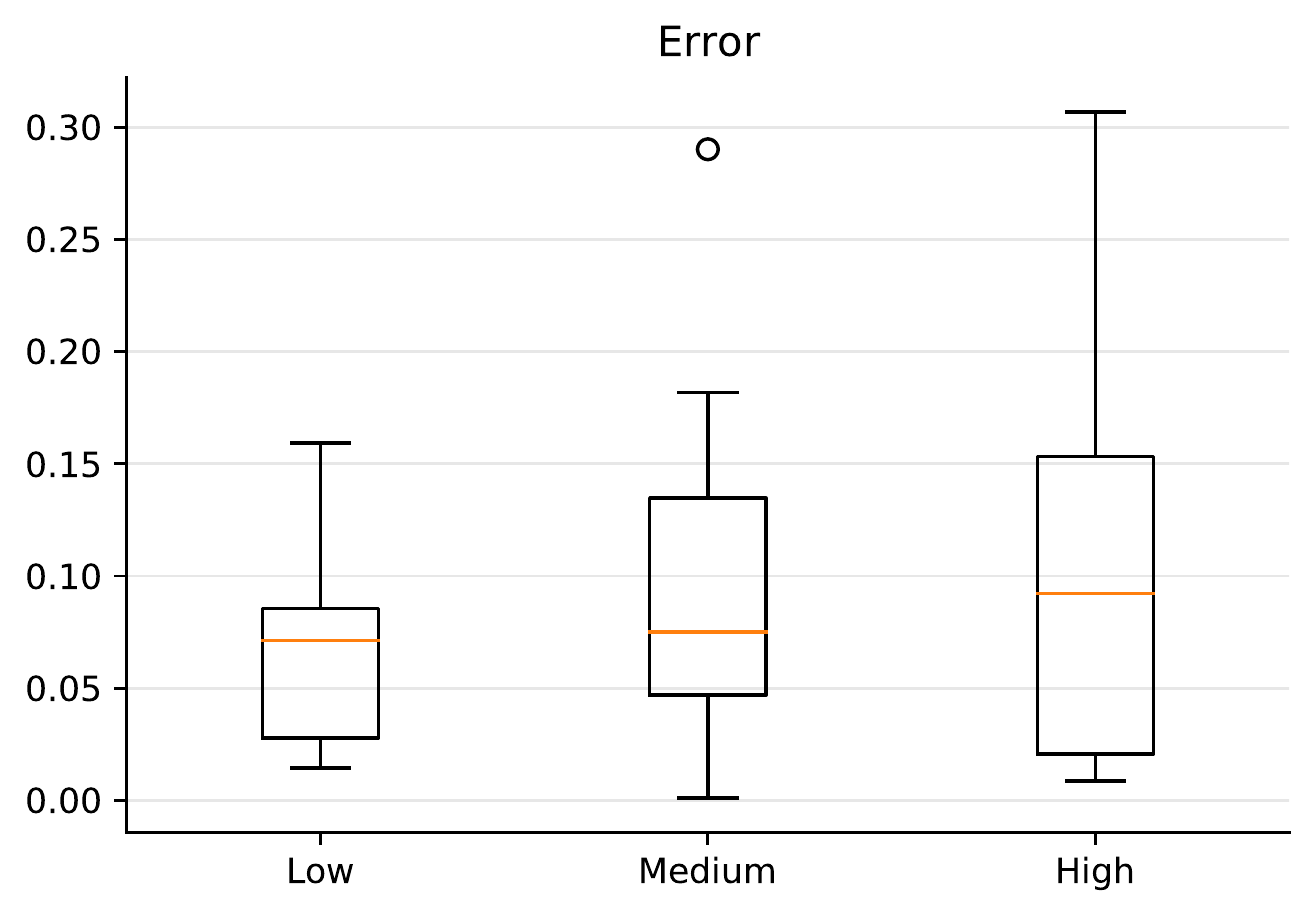}
    \caption{Fairness of the dataset. No statistically significant differences were found in either the accuracy, or the bias of the estimates across the low, medium and high performing algorithms.}
    \label{fig:fairness}
\end{figure}

\subsection{Predicting All Game Scores}
\label{sec:pred_all}

\begin{figure}[h]
    \centering
    \includegraphics[width=0.99\linewidth]{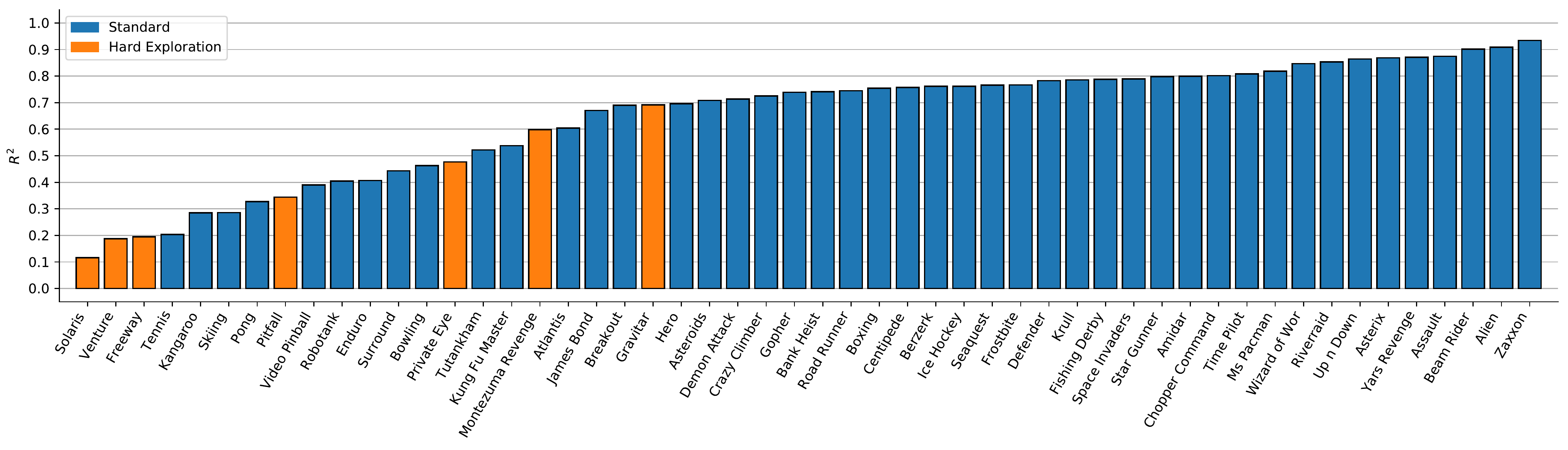}
    \caption{The five games selected in Atari-5 are able to predict accurate scores for many of the other games within the ALE benchmark. However, hard exploration exploration games are less well predicted by the subset.}
    \label{fig:other_games}
\end{figure}

We found that Atari-5 was able to produce surprisingly accurate score estimates for many of the other games in the ALE benchmark, as shown in Figure \ref{fig:other_games}. There were 17 games which showed a high degree of correlation, with \rsq greater than 0.8, and together, the models explain 71.5\% of the variance in log-scores within the dataset. 

Inspired by this result, we tested the `vital few, trivial many' principle \cite{juran2003juran},\footnote{Sometimes referred to as the 80/20 rule, or the Pareto Principle.} by repeating the experiment with the Atari-10 dataset (17.5\% of the environments) and found that those games could explain 80.0\% of the variance of the full 57-game set. Because of this, in many cases, other performance measures (for example, the number of games with human-level performance) could potentially be approximated using the Atari-10 game set. Model parameters for predicting scores for these games is given in Appendix \ref{app:coefficents}. We also note that while the Atari-5 games predict many of the games well, they perform poorly in Bellemare's hard exploration games \cite{bellemare2016unifying}. 

\section{Structure Within in the ALE}

\begin{figure}[t]
    \centering
    \includegraphics[width=0.99\linewidth]{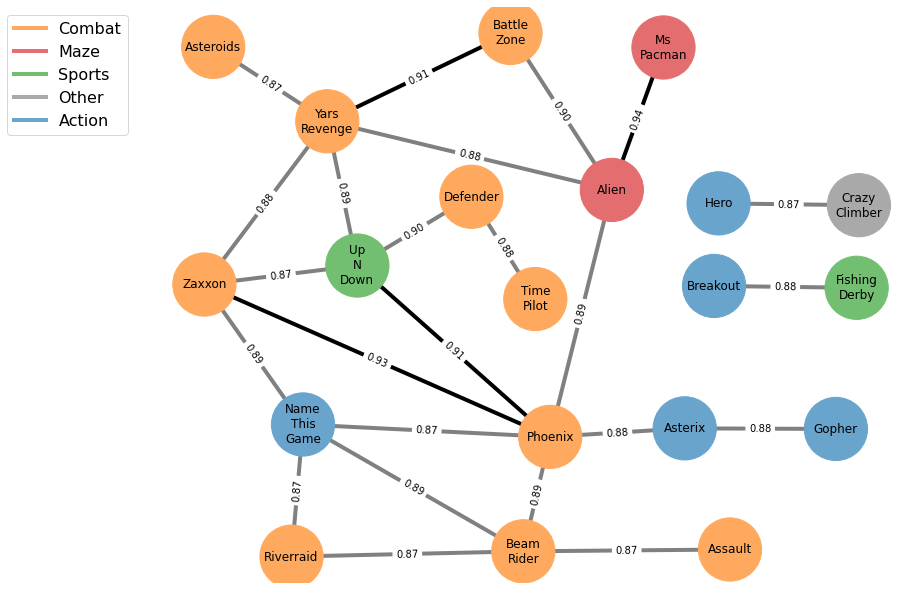}
    \caption{Structure within the ALE dataset. Edges indicate PCC, colours the games' category.}
    \label{fig:structure}
\end{figure}

While the ALE dataset contains a diverse range of games, there are still some games within the dataset that are highly correlated. Therefore, we assessed the similarity between scores in games in the ALE benchmark by calculating the Pearson Correlation Coefficient (PCC) \cite{benesty2009pearson} on the log-transformed normalized game scores for each game pair in the 57-game dataset. We considered a pair with a coefficient above 0.9 to be highly correlated. 
The most correlated twenty-four pairings are shown in Figure \ref{fig:structure}, with highly correlated pairs depicted with bold edges. We found that the two most correlated games were \textit{Alien} and \textit{Ms. Pacman}, which was not surprising as these games are both Pacman-like games with very similar rules and primarily differ only in their graphical presentation. We also separated games into categories based on their genre\footnote{Genres were taken from Wikipedia entries for each of the games.} and note that games of similar categories are often clustered together. 

We also checked to see if any games were negatively correlated and found \env{Skiing} scores were moderately negatively correlated with \env{Alien} (-0.51), \env{Frostbite} (-0.55), \env{Ms. Pacman} (-0.51), and \env{River Raid} (-0.50). \env{Skiing} is unlike most other ALE games in that all score is deferred until the terminal state of the episode and that the locally optimal but globally sub-optimal policy of racing to end ignoring penalty flags is easily learnable. This suggests that \env{Skiing} may require novel approaches to solve.

\section{Case Study}

Our experimental results show strong results for Atari-5 on the dataset used to fit the model. In addition, K-fold cross-validation scores indicate that our model is not overfitting the training data and suggest the model should generalize to unseen data well. However, changes in algorithms or environmental settings may result in poorer performance when applied to unseen data. For this reason, we performed a case study where we applied Atari-5 retroactively to a recent paper not included in our training set. We then asked the following question. 
\textit{If this paper had used Atari-5 instead of the full ALE, would the results have changed materially?} 


We set up our experiment as follows. First, we searched for accepted papers submitted to AAAI-22 that referenced ALE. We then selected the first paper,which contained results for no less than 40 games and which included results for the five games in the Atari-5 benchmark. The first paper we found meeting this requirement was \textit{Conjugated Discrete Distributions for Distributional Reinforcement Learning} \cite{lindenberg2022conjugated}, which contains results for C51, IQN, and Rainbow, as well as their own algorithm C2D.\footnote{Their results for C51, IQN, and Rainbow do not match those in our training set. This is not unexpected as results on these algorithms are highly dependant on both the implementation details, and the environmental settings.} 

We defined two evaluation metrics, which are relative error on predicted median score as compared to the true median score, and the inversion count between the order of the algorithms under the median score, as compared to the order under Atari-5. We used the arXiv preprint version of the paper, which included the supplementary material containing the required results.



The results of our study are given in Table \ref{tab:cs_abs}. Relative error on these four algorithms was higher than expected (12.6\%-26.0\%, average of 18.9\%). However, we found that Atari-5 scores consistently underpredicted the true results by a similar ratio $(\approx 0.80)$. We hypothesise that this may be due to differences in the environmental settings (evidenced by Rainbow scoring $147$ on their experiments compared to $223$ in our dataset). Since it is often useful to measure the performance of an algorithm relative to some baseline, we also considered the scores normalised to the best performing algorithm, Rainbow. We present the results in and Table \ref{tab:cs_re}. Because Atari-5 is consistent in its under prediction, with this adjustment, relative error for all models fell below 10\% (3.5\% - 9.0\%, average of 6.7\%). We note only one inversion in the order: the swapping of C51, and IQN.


\begin{table}[h]
    \centering
    \begin{tabular}{l r r r}
    \toprule
    Algorithm & Median & Atari-5 & Rel.Error \\
    \midrule
    C51        & 109 & 96 & 12.6\% \\
    IQN        & 129 & 95 & 26.0\% \\
    C2D        & 133 & 111 & 17.0\% \\
    Rainbow    & 147 & 118 & 19.8\% \\
    \bottomrule
    \end{tabular}
    \caption{Median and Atari-5 scores for the algorithms in the paper. Atari-5 consistently underestimates the true median, however the ordering of algorithms remains largely unchanged.}
    \label{tab:cs_abs}
\end{table}

\begin{table}[h]
    \centering
    \begin{tabular}{l r r r}
    \toprule
    Algorithm & Median & Atari-5 & Rel.Error \\
    \midrule
    C51        & 0.74 & 0.81 & 9.0\% \\
    IQN        & 0.87 & 0.81 & 7.7\% \\
    C2D        & 0.90 & 0.94 & 3.5\% \\
    Rainbow    & 1.00 & 1.00 & 0.0\% \\
    \bottomrule
    \end{tabular}
    \caption{Median and Atari-5 scores normalized to the Rainbow result. These results indicate the relative performance of each algorithm as compared to Rainbow.}
    \label{tab:cs_re}
\end{table}





Given these results, we conclude that, had Atari-5 been used instead of the full ALE dataset, while the median score would have varied substantially (18.9\%), the relative performance differences would have been similar (6.7\%). This degree of error was not significant enough to change the paper's outcome, which was that C2D outperformed both C51 and IQN and underperformed Rainbow. We, therefore, conclude that had Atari-5 been used, the results would not have changed materially.

\section{Discussion}

In this section, we outline the recommended use of the Atari-5 benchmark, discuss its primary value and limitations, and consider the broader impact of our work. We intend Atari-3-Val to be used for hyperparameter tuning and algorithm development and Atari-5 to be used to generate evaluation results. In some cases, researchers might want to include results for specific games that demonstrate a strength or weakness of their algorithm while also providing a measure of general performance in the form of Atari-5. Because Atari-5 and median scores differ in their bias, it is recommended to always compare Atari-5 scores with Atari-5 scores, and not against median scores directly. For this reason, we have provided Atari-5 scores for common algorithms in Table \ref{tab:popular}. 


Atari-5 provides three key points of value. First, it captures a significant portion of the information gained from a full ALE evaluation using one-tenth of the evaluations. Second, it provides a standard subset to ALE, selected using a principled process. Third, it provides backwards compatibility with prior work allowing results on previous algorithms to be generated after the fact, so long as scores for the five games were included.

\subsection{Limitations}
\label{sec:limitations}

Algorithms included in our training dataset are very diverse and include neural network and non-neural network approaches. Because of this, Atari-5 generates results robust to \textit{algorithmic} and \textit{training} choices, such as how many frames an algorithm was trained for. However, any changes made to the \textit{environment}, importantly the maximum number of frames, could affect the result, as they modify the optimal policy and potentially the maximum score. We also note that stochasticity in the form of the probability of repeating actions is an environmental change but that our dataset contains both stochastic and deterministic variants. This does not seem to make a significant difference to the prediction accuracy.

We also note that Atari-5 does not do well at incorporating hard-exploration games (see Figure \ref{fig:other_games}). We, therefore, recommend, for algorithms where this matters, to run individual tests for these games. Researchers could then establish their algorithm's \textit{general} performance via Atari-5, while also demonstrating a \textit{specific} ability on  hard-exploration problems by giving individual results on those games.

\subsection{Broader Impact}

We believe that Atari-5 plays a vital role in increasing access and diversity in RL research, thus increasing reproducibility. Our work aims to address several issues that currently affect RL researchers. One of the barriers to working in RL research is the high level of computation required to produce results on common benchmarks. Researchers who do not have access to tens of thousands of dollars of compute capacity cannot evaluate their algorithms on established benchmarks, inhibiting their ability to publish their work. This has led to a lack of diversity in the literature, evidenced by the fact that only two companies have been able to produce results on the full dataset in six years. We believe Atari-5 is an important step in the right direction, as it brings Atari games inside the sphere of computationally accessible benchmarks. 




\section{Conclusions}

We have introduced a new benchmark, Atari-5, that provides median score estimates to within 10\% of their true values but at less than one-tenth the cost. A key feature of Atari-5 is that, as a subset of ALE, it can be applied retroactively to prior results, without retraining, so long as scores for all five games are provided. We have also shown that our extended benchmark \textit{Atari-10} can generate useful score estimates for \textit{all} games within the ALE dataset. We hope this new benchmark can offset the trend of increasingly high computation requirements required to develop and reproduce results on the ALE benchmark and increase RL participation while providing backwards compatible results to prior work.

\clearpage

\bibliography{references}

\newpage

\appendix

\onecolumn

\section{Estimation of Cost to Reproduce for Selected Papers}
\label{app:cost_estimate}

Retroactive estimation of costs and carbon emissions can be difficult \cite{patterson2021carbon}. Due to missing details in the publications, our estimations required many assumptions about the hardware and time needed to reproduce a result. Our estimates do not include overheads such as air-conditioning, testing, hyper-parameter searches, or ablation studies. Therefore these estimates should be taken as useful \textit{lower bound} estimates on the cost to reproduce a work. We also note the large difference between preemptive and standard costs which are approximately one-third of the cost. However, preemptive computing may not always be feasible due to congestion near conference deadlines. We took the computing requirements from the papers and used Google Cloud Compute to estimate the cost of reproducing the work.\footnote{\url{https://cloud.google.com/compute/gpus-pricing}}

For some older algorithms, it may be possible to reduce the costs given here by training on more modern hardware (for example, an A100 or TPU). However, if the algorithm is CPU constrained, this could increase, not decrease, the cost due to the underutilization of the device. Therefore we estimate reproduction costs with our best guess at the hardware that would have been used at the time of publication. A summary of the calculations and results is given in Table \ref{tab:cost_calculations}. In addition, we detail the assumptions made for each algorithm in the paragraphs below.

\begin{table*}[h]
    \centering
    \caption{Estimated cost to reproduce a full evaluation on ALE. These are the estimated costs for a \textit{single} run, some papers include multiple runs, or ablation studies.}
    \begin{tabular}{l l l l l l}
    \toprule
        Year & Algorithm & Compute / env. & GPU hours & Cost (preemptive) & Cost (standard) \\
    \midrule
        2018 & Rainbow DQN \cite{hessel2018rainbow} & 10 GPU days & 13,680 & \$10,670 USD & \$35,759 USD \\
        2020 & Agent-57 \cite{badia2020agent57} & 375 hours (est) & 21,375 & \$30,673 USD & \$110,352 USD \\
    \bottomrule
    \end{tabular}
    \label{tab:cost_calculations}
\end{table*}

\paragraph{Rainbow DQN}
Rainbow DQN required 10 days on a single GPU for each of the 57 environments.\cite{hessel2018rainbow} The hardware used is not mentioned in the paper so we assume a V100, with an E2-standard-4 (16GB) instance,\footnote{Smaller instances might not be feasible due to RAM requirements.} which has a GPU/CPU cost per hour of \$2.48/\$0.134 (standard), and \$0.74/\$0.04 (preemptive). We also note that \cite{revisiting2020} also produced estimates for Rainbow DQN, but used 5-days of GPU time instead of 10.

\paragraph{Agent-57}
Agent-57 trains at $\approx 260$ environment steps per second per actor with 256 actors \cite{badia2020agent57}. To generate the 90B frames needed would then take 375 hours. The paper states that a single GPU is used, and for the purposes of our estimate, we assume that this also requires 375 hours.\footnote{Even though Agent-57 uses a distributed training/roll-out architecture, we believe this to be reasonable based on the assumption that actors and learners are limited in how out of sync they can become.} The hardware used is not stated, so we assume an A100, which is the best available GPU at the time of their publication. We assume a 64 CPU machine (\textit{N2-highcpu-64}) would be needed as our observation is that Atari environments typically run at 1,000 environment steps per CPU. The GPU/CPU cost per hour is \$2.9333/\$2.2294 (standard) and \$0.88/\$0.555 (preemptive).

\paragraph{Power Usage and Carbon Emissions}
We estimate the power required to reproduce the Agent-57 result as follows. A single A100 uses up to 400W.\footnote{NVIDIA lists two variants at \url{https://www.nvidia.com/en-au/data-center/a100/} one with a 400W max TDP and one with 250W. Cards likely run below their maximum power limits. However, this is partly balanced out by power supply inefficiencies.} We use 215W as the estimated power consumption for the CPU, which is taken from the Xeon Phi 7210 specifications.\footnote{\url{https://ark.intel.com/content/www/us/en/ark/products/94033/intel-xeon-phi-processor-7210-16gb-1-30-ghz-64-core.html}} We do not include estimates for air-conditioning, power supply inefficiency, or other components such as RAM, networking, and storage. The use of TDP is only an approximate measure as devices frequently run under the TDP limit and may also occasionally run over it too. Using the estimates above, the total power required to reproduce a result would be 13.166mWh. Because our compute estimations were based on Google Cloud Compute servers in the U.S., we used the 2019 U.S. carbon emissions rate of 0.92 pounds per kWh.\footnote{\url{https://www.eia.gov/tools/faqs/faq.php?id=74&t=11}} We note that this could be higher in developing countries. This equates to approximately 6.05 tones of CO2 per 57-game run. 

\section{Subsets Used by Other Papers}
\label{app:subsets}

\begin{table}[h]
    \centering
    \begin{tabular}{l l}
    \toprule
         Paper & Games  \\
    \midrule
         DQN \cite{mnih2013playing} & Beam Rider, Breakout, Enduro, Pong,\\ & Q*bert, Seaquest, Space Invaders. \\ 
         \specialrule{0pt}{1.5pt}{1.5pt}
         A3C \cite{mnih2016asynchronous} & Beam Rider, Breakout, Pong, Q*bert, Space Invaders. \\
         \specialrule{0pt}{1.5pt}{1.5pt}
         Unifying Count-Based Exploration (Hard Exploration) & Freeway, Gravitar, Montezuma's Revenge, Pitfall!, \\ \cite{bellemare2016unifying} & Private Eye, Solaris, Venture. \\
         \specialrule{0pt}{1.5pt}{1.5pt}
         Playing hard exploration games (Hard Exploration) & Montezuma's Revenge, Pitfall, Private Eye. \\
         \cite{aytar2018playing} & \\
         \specialrule{0pt}{1.5pt}{1.5pt}
         Agent57 (Challenging Set) & Freeway, Gravitar, Montezuma's Revenge, Pitfall! \\ \cite{badia2020agent57} & Private Eye, Solaris Venture, Beam Rider, Pong, Skiiing. \\
        \specialrule{0pt}{1.5pt}{1.5pt}
         Compress and Control \cite{veness2015compress} & Q*bert, Pong, Freeway.\\
    \bottomrule
    \end{tabular}
    \caption{Subsets of Atari used by various works.}
    \label{tab:subsets}
\end{table}

\section{Proof of Proposition 2}
\label{app:monotonic}

We prove the more general case of Proposition 2 using some fixed (weakly) increasing $\phi: \mathbb{R} \rightarrow \mathbb{R}$, where Proposition 2 can be recovered by selecting $\phi(x)=x$.

Consider two algorithms $a, b$, and a set of scores for each of $n$ environments $a_1...a_n$, $b_1...b_n$, where $a_i$ is the score on the $i$-th environment for algorithm $a$, and $b_i$ is the score on the $i$-th environment for algorithm $b$. Given some (weakly) increasing function $\phi: \mathbb{R} \rightarrow \mathbb{R}$, and constants $c_i \in \mathbb{R} \geq 0$, we have that
\begin{align}
    &a_i \geq b_i & \quad \forall \; i \in [1..n] \\
    \Rightarrow & \phi(a_i) \geq \phi(b_i) & (\phi \text{ is weakly increasing}) \\
    \Rightarrow & c_i \phi(a_i) \geq c_i \phi(b_i) & (\text{for } c_i \geq 0). \\
\end{align}

Since this holds for all $i \in [1..n]$ we have 

\begin{align}
\sum_{i=1}^n c_i \phi(a_i) \geq \sum_{i=1}^n c_i \phi(b_i) 
\end{align}

as required.

\section{Algorithms Included in the Dataset}
\label{app:algo}

We included in our dataset only algorithms with results on 40 or more ALE games. They are listed in Table \ref{tab:dataset_algorithms}, along with how many games each algorithm had results for. Games that are not part of the 57-game canonical set were not included. The names are provided from PapersWithCode, and some algorithms appear multiple times as they included multiple configurations.

\begin{table}[h]
    \caption{Algorithms included in the dataset.}
    \label{tab:dataset_algorithms}
    
    \centering
    \footnotesize
    \begin{tabular}{l r r}
    \toprule
    Algorithm & Games & Median Score \\
    \midrule
    MuZero                                   & 57   & 2041.1    \\
    Agent57                                  & 57   & 1975.8    \\
    R2D2                                     & 57   & 1926.7    \\
    SEED 8 TPU v3 cores                      & 57   & 1846.5    \\
    R2D2(Retrace)                            & 57   & 1342.8    \\
    NGU (32-0)                               & 57   & 1208.1    \\
    GDI-H3(200M frames)                      & 56   & 1030.9    \\
    MuZero (Res2 Adam)                       & 57   & 1010.6    \\
    GDI-I3                                   & 50   & 868.0     \\
    NGU (32-0.3)                             & 57   & 568.8     \\
    LASER No Sweep (200M)                    & 57   & 454.9     \\
    Ape-X                                    & 57   & 434.1     \\
    LASER Shared (200M)                      & 57   & 413.7     \\
    LASER Shared (50M)                       & 57   & 365.0     \\
    IQN                                      & 57   & 237.8     \\
    Rainbow (noop)                           & 56   & 227.0     \\
    DreamerV2                                & 55   & 214.7     \\
    QR-DQN-1                                 & 57   & 210.7     \\
    QR-DQN-0                                 & 57   & 199.2     \\
    Prior+Duel noop                          & 50   & 194.8     \\
    IMPALA (deep)                            & 57   & 191.8     \\
    Reactor                                  & 57   & 186.8     \\
    Planning                                 & 50   & 183.3     \\
    Reactor ND                               & 57   & 180.4     \\
    C51                                      & 57   & 177.7     \\
    NoisyNet-Dueling                         & 54   & 174.4     \\
    Bootstrapped DQN                         & 49   & 153.4     \\
    DDQN+Pop-Art noop                        & 49   & 147.4     \\
    Distrib DQN (noop)                       & 56   & 146.7     \\
    Duel DDQN (noop)                         & 56   & 142.3     \\
    A2C + SIL                                & 49   & 140.0     \\
    Prior noop                               & 49   & 140.0     \\
    Priority DDQN (noop)                     & 56   & 136.6     \\
    A3C LSTM hs                              & 50   & 133.2     \\
    Duel hs                                  & 50   & 131.4     \\
    Prior+Duel hs                            & 52   & 128.8     \\
    A3C FF hs                                & 50   & 120.0     \\
    Prior hs                                 & 50   & 118.0     \\
    Persistent AL                            & 57   & 116.5     \\
    DDQN (noop)                              & 56   & 115.2     \\
    Noisy DQN (noop)                         & 56   & 114.9     \\
    Advantage Learning                       & 57   & 114.3     \\
    DDQN (tuned) hs                          & 50   & 114.3     \\
    Human Ref                                & 57   & 100.0     \\
    Adam Huber                               & 54   & 97.8      \\
    Adam MSE                                 & 54   & 92.5      \\
    Bellman                                  & 57   & 87.7      \\
    NEAT Object                              & 56   & 87.2      \\
    POP3D                                    & 49   & 83.6      \\
    Nature DQN                               & 48   & 82.7      \\
    DQN (noop)                               & 56   & 77.6      \\
    CMA-ES Object                            & 56   & 77.5      \\
    H-NEAT Object                            & 56   & 76.7      \\
    DQN hs                                   & 50   & 70.7      \\
    CNE Object                               & 56   & 70.5      \\
    NEAT Noise                               & 56   & 69.2      \\
    A3C FF (1 day) hs                        & 50   & 68.2      \\
    Gorila                                   & 49   & 68.1      \\
    CNE Noise                                & 56   & 58.0      \\
    RMSProp Huber                            & 54   & 57.3      \\
    ES FF (1 hour) noop                      & 45   & 55.8      \\
    CGP                                      & 56   & 53.0      \\
    \bottomrule
    \end{tabular}
    
\end{table}

\section{Models for Individual Games}
\label{app:coefficents}

We provide coefficients for each of the Atari-X and Atari-Val-X models in Table \ref{tab:coefficents}, as well as models for predicting scores for all games within the 57-game ALE dataset (Table \ref{tab:game_models_5}, \ref{tab:game_models_10}). An estimated normalized game score ($s_\text{est}$) can be calculated using the following formula

\begin{align}
s_\text{log} &= c + \sum_{i=1}^{10} x_i \log_{10}(1+s_i) \\
s_\text{est} &= 10^{s_\text{log}} - 1
\end{align}

where $s_i$ are the normalized scores for the appropriate game using normalization constants taken from Table \ref{app:norm_constants}. Not all models make use of the intercept constant $c$.

\begin{table*}[h]
    \centering
    \footnotesize
    \caption{Parameters for each of the best subset models}
    \begin{tabular}{l l l}
        \toprule
        Dataset & Games & Parameters \\
        \midrule
        Atari-1 & NameThisGame & $0.9976$  \vspace{0.2em}\\
        Atari-3 & BattleZone, NameThisGame, Phoenix & $0.3706, 0.5133, 0.1015$ \vspace{0.2em}\\
        Atari-5 & BattleZone, DoubleDunk, NameThisGame, Phoenix, Qbert & $0.3820, 0.0679, 0.3108, 0.1241, 0.0805 $ \vspace{0.2em}\\
        
        Atari-10 & Amidar, Bowling, Frostbite, KungFuMaster, RiverRaid & $0.0825, 0.0559, 0.0691, 0.0986, 0.0486,$ \\
        & BattleZone, DoubleDunk, NameThisGame, Phoenix, QBert & $0.1888, 0.0852, 0.1287, 0.1643, 0.0592$  \vspace{0.2em}\\
        
        Atari-3-Val & Assault, MsPacMan, YarsRevenge & $0.3353, 0.4236, 0.1916$ \vspace{0.2em}\\
        Atari-5-Val & BankHeist, VideoPinball, Assault, MsPacman, YarsRevenge & $0.1072, 0.0959, 0.2234, 0.2943, 0.2239$ \\
        \bottomrule
    \end{tabular}
    \label{tab:coefficents}
\end{table*}

\begin{table*}[h]
\footnotesize
    \centering
    \caption{Atari-5 linear regression models for each game in the ALE.}
    \begin{tabular}{l r r r r r r}
    \toprule
        Game & c & $x_1$ & $x_2$ & $x_3$ & $x_4$ & $x_5$ \\
    \midrule
        Alien & -0.807 & 0.717 & -0.106 & 0.362 & 0.195 & 0.100 \\
        Amidar & -0.180 & 0.426 & -0.013 & 0.289 & -0.160 & 0.471 \\
        Assault & -0.559 & 0.015 & 0.389 & 0.669 & 0.191 & 0.204 \\
        Asterix & -0.449 & -0.478 & -0.149 & 0.897 & 0.665 & 0.356 \\
        Asteroids & -2.150 & 0.671 & 0.114 & -0.214 & 0.791 & -0.043 \\
        Atlantis & 1.860 & -0.097 & 0.134 & -0.016 & 0.245 & 0.346 \\
        Bankheist & -0.342 & 0.378 & 0.095 & 0.447 & -0.128 & 0.334 \\
        Battlezone & 0.000 & 1.000 & 0.000 & 0.000 & 0.000 & 0.000 \\
        Beamrider & -1.113 & -0.152 & -0.045 & 1.149 & 0.259 & 0.100 \\
        Berzerk & -0.556 & 0.472 & 0.065 & -0.058 & 0.568 & -0.038 \\
        Bowling & 0.772 & 0.848 & -0.180 & 0.574 & -0.415 & -0.301 \\
        Boxing & 1.581 & -0.138 & 0.396 & 0.088 & 0.023 & 0.050 \\
        Breakout & 0.998 & -0.574 & -0.017 & 0.873 & 0.134 & 0.398 \\
        Centipede & -1.313 & 1.013 & -0.161 & 0.642 & 0.550 & -0.599 \\
        Choppercommand & -0.988 & 1.051 & 0.065 & -0.461 & 0.595 & 0.114 \\
        Crazyclimber & 1.105 & -0.208 & -0.048 & 0.425 & 0.015 & 0.472 \\
        Defender & 0.584 & 0.486 & 0.234 & -0.818 & 0.699 & 0.095 \\
        Demonattack & 1.052 & -0.380 & 0.076 & 0.449 & 0.392 & 0.338 \\
        Doubledunk & 0.000 & 0.000 & 1.000 & 0.000 & 0.000 & 0.000 \\
        Enduro & 0.100 & 0.833 & -0.064 & 0.918 & -1.163 & 0.575 \\
        Fishingderby & 1.412 & -0.119 & 0.094 & 0.224 & 0.030 & 0.119 \\
        Freeway & 1.297 & 0.713 & 0.020 & 0.125 & -0.439 & -0.060 \\
        Frostbite & -1.273 & 0.927 & -0.229 & 1.166 & -0.327 & 0.073 \\
        Gopher & 0.597 & -0.226 & 0.035 & 0.665 & 0.301 & 0.221 \\
        Gravitar & -0.408 & 1.351 & 0.133 & -0.561 & -0.035 & -0.025 \\
        Hero & 0.927 & -0.118 & -0.107 & 0.286 & 0.070 & 0.277 \\
        Icehockey & 0.152 & 0.363 & 0.234 & 0.197 & 0.092 & -0.105 \\
        Jamesbond & -0.133 & 0.881 & 0.320 & -0.226 & 0.440 & -0.139 \\
        Kangaroo & 1.482 & 0.209 & -0.089 & 0.546 & -0.245 & 0.090 \\
        Krull & 0.201 & 0.999 & 0.194 & 0.205 & -0.126 & -0.020 \\
        Kungfumaster & 1.339 & 0.270 & 0.049 & -0.055 & 0.182 & -0.022 \\
        Montezumarevenge & -0.070 & 1.256 & -0.178 & -1.161 & 0.118 & 0.316 \\
        Mspacman & -0.118 & 0.561 & -0.113 & 0.684 & -0.137 & -0.012 \\
        Namethisgame & 0.000 & 0.000 & 0.000 & 1.000 & 0.000 & 0.000 \\
        Phoenix & 0.000 & 0.000 & 0.000 & 0.000 & 1.000 & 0.000 \\
        Pitfall & 0.891 & 0.453 & -0.188 & -0.391 & -0.038 & 0.171 \\
        Pong & 1.401 & -0.168 & 0.085 & 0.200 & -0.015 & 0.128 \\
        Privateeye & 1.016 & 1.326 & -0.226 & -0.855 & -0.194 & -0.042 \\
        Qbert & 0.000 & 0.000 & 0.000 & 0.000 & 0.000 & 1.000 \\
        Riverraid & 0.125 & -0.167 & -0.161 & 1.025 & 0.053 & 0.072 \\
        Roadrunner & 0.385 & 0.180 & 0.093 & 0.600 & -0.039 & 0.234 \\
        Robotank & 1.262 & 0.036 & 0.178 & 0.490 & -0.118 & 0.010 \\
        Seaquest & -2.222 & 1.123 & -0.235 & 0.900 & -0.226 & 0.263 \\
        Skiing & 1.864 & -0.083 & 0.400 & -0.518 & -0.163 & 0.080 \\
        Solaris & 1.019 & 0.435 & 0.376 & -0.844 & 0.004 & 0.040 \\
        Spaceinvaders & 0.171 & -0.170 & 0.168 & 0.120 & 0.591 & 0.176 \\
        Stargunner & 0.676 & -0.123 & -0.056 & 0.380 & 0.423 & 0.235 \\
        Surround & 0.686 & -0.290 & 0.138 & 0.744 & -0.068 & -0.059 \\
        Tennis & 1.637 & 0.123 & 0.219 & -0.344 & 0.059 & 0.140 \\
        Timepilot & -0.729 & 0.889 & 0.281 & -0.406 & 0.688 & -0.068 \\
        Tutankham & 0.357 & -0.035 & 0.067 & 0.776 & -0.181 & 0.143 \\
        Upndown & -0.351 & 0.215 & 0.166 & -0.110 & 0.596 & 0.267 \\
        Venture & 0.415 & 0.946 & -0.102 & -0.008 & -0.611 & 0.336 \\
        Videopinball & 1.846 & 0.021 & 0.114 & 0.212 & 0.165 & 0.052 \\
        Wizardofwor & 0.104 & 0.503 & 0.082 & 0.229 & 0.188 & 0.008 \\
        Yarsrevenge & -0.082 & 0.823 & 0.074 & -0.494 & 0.342 & 0.121 \\
        Zaxxon & -0.145 & 0.355 & -0.060 & 0.468 & 0.250 & 0.029 \\
    \bottomrule
    \end{tabular}
\label{tab:game_models_5}
\end{table*}

\begin{table*}[h]
\footnotesize
    \centering
    \caption{Atari-10 linear regression models for each game in the ALE.}
    \resizebox{\textwidth}{!}{%
    \begin{tabular}{l r r r r r r r r r r r}
    \toprule
        Game & c & $x_1$ & $x_2$ & $x_3$ & $x_4$ & $x_5$ & $x_6$ & $x_7$ & $x_8$ & $x_9$ & $x_{10}$ \\
    \midrule
        Alien & -1.411 & 0.457 & 0.224 & -0.024 & 0.336 & 0.512 & 0.337 & -0.006 & -0.354 & 0.245 & -0.065 \\
        Amidar & 0.000 & 1.000 & 0.000 & 0.000 & 0.000 & 0.000 & 0.000 & 0.000 & 0.000 & 0.000 & 0.000 \\
        Assault & -0.094 & 0.124 & -0.167 & 0.152 & 0.070 & -0.463 & -0.237 & 0.290 & 1.069 & 0.195 & 0.151 \\
        Asterix & 0.246 & 0.209 & -0.063 & 0.354 & 0.110 & -0.324 & -1.084 & -0.186 & 0.898 & 0.732 & 0.302 \\
        Asteroids & -3.862 & -0.270 & 0.260 & -0.454 & 0.606 & 0.998 & 0.900 & 0.197 & -0.881 & 0.708 & 0.102 \\
        Atlantis & 2.692 & 0.128 & -0.352 & 0.180 & -0.043 & -0.413 & -0.209 & 0.015 & 0.405 & 0.202 & 0.234 \\
        Bankheist & 0.675 & -0.072 & 0.165 & 0.293 & -0.714 & -0.476 & 0.238 & 0.199 & 0.360 & 0.265 & 0.349 \\
        Battlezone & 0.000 & 0.000 & 0.000 & 0.000 & 0.000 & 0.000 & 1.000 & 0.000 & 0.000 & 0.000 & 0.000 \\
        Beamrider & -0.606 & 0.146 & -0.011 & 0.059 & -0.187 & 0.199 & -0.287 & -0.021 & 0.875 & 0.308 & 0.040 \\
        Berzerk & -1.226 & 0.343 & -0.289 & -0.211 & 0.519 & -0.297 & 0.590 & -0.097 & 0.570 & 0.371 & -0.247 \\
        Bowling & 0.000 & 0.000 & 1.000 & 0.000 & 0.000 & 0.000 & 0.000 & 0.000 & 0.000 & 0.000 & 0.000 \\
        Boxing & 1.983 & 0.067 & -0.067 & 0.079 & -0.158 & -0.216 & -0.181 & 0.378 & 0.217 & 0.085 & 0.001 \\
        Breakout & 1.858 & 0.339 & -0.264 & 0.090 & -0.229 & -0.106 & -0.626 & -0.079 & 0.992 & 0.108 & 0.198 \\
        Centipede & -3.331 & -0.319 & 0.412 & -0.216 & 0.956 & -0.241 & 0.763 & -0.199 & 0.978 & 0.492 & -0.307 \\
        Choppercommand & -3.172 & 0.415 & -0.136 & -0.621 & 1.290 & 0.548 & 1.185 & -0.118 & -0.159 & 0.064 & -0.027 \\
        Crazyclimber & 1.689 & 0.382 & -0.153 & 0.085 & -0.120 & 0.152 & -0.353 & -0.049 & 0.194 & 0.027 & 0.261 \\
        Defender & 0.160 & -0.295 & -0.080 & -0.104 & 0.141 & 0.694 & 0.876 & 0.312 & -1.328 & 0.562 & 0.149 \\
        Demonattack & 1.261 & 0.137 & -0.124 & 0.051 & 0.219 & 0.059 & -0.633 & 0.005 & 0.474 & 0.294 & 0.304 \\
        Doubledunk & 0.000 & 0.000 & 0.000 & 0.000 & 0.000 & 0.000 & 0.000 & 1.000 & 0.000 & 0.000 & 0.000 \\
        Enduro & 2.754 & 0.412 & -0.137 & 0.809 & -0.718 & -0.384 & -0.092 & -0.017 & 0.503 & -0.884 & 0.412 \\
        Fishingderby & 1.736 & 0.098 & -0.038 & 0.041 & -0.154 & 0.018 & -0.141 & 0.106 & 0.137 & 0.084 & 0.054 \\
        Freeway & 3.306 & -0.284 & -0.419 & 0.338 & -1.026 & -1.388 & 1.052 & -0.102 & 1.292 & -0.196 & -0.068 \\
        Frostbite & 0.000 & 0.000 & 0.000 & 1.000 & 0.000 & 0.000 & 0.000 & 0.000 & 0.000 & 0.000 & 0.000 \\
        Gopher & 0.717 & -0.030 & -0.297 & 0.228 & 0.632 & -0.626 & -0.680 & -0.169 & 1.380 & 0.107 & 0.273 \\
        Gravitar & -1.119 & -0.132 & 0.272 & 0.080 & 0.368 & 0.182 & 1.054 & 0.229 & -1.010 & 0.072 & 0.079 \\
        Hero & 1.319 & 0.255 & 0.042 & 0.051 & -0.237 & 0.440 & -0.209 & -0.010 & -0.332 & 0.171 & 0.137 \\
        Icehockey & -0.103 & -0.174 & -0.087 & -0.130 & 0.028 & 0.022 & 0.675 & 0.208 & 0.382 & 0.013 & -0.066 \\
        Jamesbond & -0.386 & -0.222 & -0.565 & 0.132 & 0.817 & -0.212 & 0.916 & 0.116 & 0.380 & 0.009 & -0.107 \\
        Kangaroo & 1.989 & 0.658 & 0.088 & 0.306 & 0.132 & -0.179 & -0.625 & -0.075 & 0.235 & -0.076 & -0.143 \\
        Krull & 0.858 & 0.043 & 0.163 & 0.392 & 0.016 & -0.588 & 0.251 & 0.171 & 0.355 & 0.030 & 0.084 \\
        Kungfumaster & 0.000 & 0.000 & 0.000 & 0.000 & 1.000 & 0.000 & 0.000 & 0.000 & 0.000 & 0.000 & 0.000 \\
        Montezumarevenge & 0.012 & -0.465 & 0.244 & 0.338 & 0.069 & 0.189 & 0.922 & -0.043 & -1.737 & 0.226 & 0.581 \\
        Mspacman & -0.492 & 0.150 & 0.255 & 0.031 & 0.103 & 0.363 & 0.323 & 0.007 & 0.072 & -0.026 & -0.047 \\
        Namethisgame & 0.000 & 0.000 & 0.000 & 0.000 & 0.000 & 0.000 & 0.000 & 0.000 & 1.000 & 0.000 & 0.000 \\
        Phoenix & 0.000 & 0.000 & 0.000 & 0.000 & 0.000 & 0.000 & 0.000 & 0.000 & 0.000 & 1.000 & 0.000 \\
        Pitfall & 0.523 & 0.270 & 0.606 & 0.170 & 0.033 & 0.710 & -0.198 & 0.086 & -1.748 & 0.238 & 0.182 \\
        Pong & 1.846 & 0.191 & -0.012 & 0.092 & -0.190 & 0.163 & -0.276 & 0.133 & -0.115 & 0.058 & 0.023 \\
        Privateeye & 0.063 & 0.067 & 0.889 & -0.111 & -0.042 & 1.267 & 0.919 & 0.112 & -2.498 & 0.015 & 0.124 \\
        Qbert & 0.000 & 0.000 & 0.000 & 0.000 & 0.000 & 0.000 & 0.000 & 0.000 & 0.000 & 0.000 & 1.000 \\
        Riverraid & 0.000 & 0.000 & 0.000 & 0.000 & 0.000 & 1.000 & 0.000 & 0.000 & 0.000 & 0.000 & 0.000 \\
        Roadrunner & 1.603 & 0.024 & -0.218 & 0.329 & -0.302 & -0.221 & -0.046 & 0.079 & 0.558 & 0.076 & 0.170 \\
        Robotank & 1.923 & 0.187 & -0.308 & 0.050 & -0.184 & -0.276 & 0.147 & 0.093 & 0.844 & -0.161 & -0.148 \\
        Seaquest & -1.653 & 0.450 & 0.214 & -0.055 & -0.608 & 1.048 & 1.125 & -0.016 & -0.360 & -0.061 & 0.047 \\
        Skiing & 1.881 & 0.124 & 0.103 & -0.442 & -0.382 & -0.298 & 0.183 & 0.276 & 0.207 & -0.176 & 0.125 \\
        Solaris & -0.063 & -0.084 & 0.642 & -0.172 & 0.331 & 0.213 & -0.049 & 0.453 & -1.175 & 0.130 & 0.305 \\
        Spaceinvaders & -0.066 & 0.502 & -0.173 & -0.152 & 0.455 & -0.141 & -0.434 & -0.016 & 0.596 & 0.381 & -0.006 \\
        Stargunner & 1.104 & 0.150 & -0.328 & -0.182 & -0.271 & 0.114 & 0.326 & -0.126 & 0.593 & 0.316 & 0.065 \\
        Surround & 0.427 & 0.133 & -0.037 & -0.076 & 0.087 & 0.245 & -0.326 & 0.126 & 0.643 & -0.117 & -0.074 \\
        Tennis & 1.574 & -0.010 & -0.087 & 0.090 & 0.274 & -0.208 & -0.047 & 0.156 & -0.120 & -0.020 & 0.160 \\
        Timepilot & -1.986 & -0.348 & 0.322 & -0.112 & 0.524 & 0.471 & 0.808 & 0.376 & -0.892 & 0.688 & 0.149 \\
        Tutankham & 0.340 & 0.203 & 0.167 & -0.006 & 0.004 & 0.059 & -0.303 & 0.088 & 0.612 & -0.112 & 0.119 \\
        Upndown & -0.761 & -0.248 & -0.224 & 0.011 & 0.493 & 0.132 & 0.296 & 0.104 & -0.031 & 0.399 & 0.332 \\
        Venture & 1.629 & -0.086 & 0.422 & 0.308 & -0.870 & 0.428 & 0.602 & 0.150 & -1.086 & -0.186 & 0.433 \\
        Videopinball & 1.715 & -0.120 & 0.007 & -0.035 & 0.098 & -0.465 & -0.037 & 0.031 & 0.713 & 0.207 & 0.134 \\
        Wizardofwor & -0.557 & -0.098 & 0.035 & 0.020 & 0.487 & 0.006 & 0.360 & 0.066 & 0.242 & 0.099 & 0.077 \\
        Yarsrevenge & -0.154 & -0.186 & -0.012 & 0.076 & 0.122 & 0.116 & 0.789 & 0.097 & -0.667 & 0.362 & 0.177 \\
        Zaxxon & -0.188 & -0.035 & 0.097 & 0.014 & -0.053 & 0.538 & 0.365 & 0.048 & -0.155 & 0.280 & 0.034 \\
    \bottomrule
    \end{tabular}
    }
    \label{tab:game_models_10}
\end{table*}

\section{Normalization Constants}

For completeness we include the average human reference scores used for normalization in Table \ref{tab:norm_scores} which have been taken from \cite{badia2020agent57}. Note, these differ from the professional human scores given by \cite{mnih2015human}. 

\begin{table}[h]
    \footnotesize
    \centering
    \caption{The canonical 57 Atari games with random and human scores used for normalization.}
    \label{app:norm_constants}

    \label{tab:norm_scores}
    
    \begin{tabular}{l r r}
    \toprule
    Game     & Random & Av. Human \\
    \midrule
    Alien    & 227.75 & 7127.7 \\
    Amidar   & 5.77 & 1719.5 \\
    Assault  & 222.39 & 742.0 \\
    Asterix  & 210.0 & 8503.3 \\
    Asteroids & 719.1 & 47388.7 \\
    Atlantis & 12850.0 & 29028.1 \\
    Bank Heist & 14.2 & 753.1 \\
    Battle Zone & 2360.0 & 37187.5 \\
    Beam Rider & 363.88 & 16926.5 \\
    Berzerk & 123.65 & 2630.4 \\
    Bowling & 23.11 & 160.7 \\
    Boxing & 0.05 & 12.1 \\
    Breakout & 1.72 & 30.5 \\
    Centipede & 2090.87 & 12017.0 \\
    Chopper Command & 811.0 & 7387.8 \\
    Crazy Climber & 10780.5 & 35829.4 \\
    Defender & 2874.5 & 18688.9 \\
    Demon Attack & 152.07 & 1971.0 \\
    DoubleDdunk & -18.55 & -16.4 \\
    Enduro & 0.0 & 860.5 \\
    Fishing Derby & -91.71 & -38.7 \\
    Freeway & 0.01 & 29.6 \\
    Frostbite & 65.2 & 4334.7 \\
    Gopher & 257.6 & 2412.5 \\
    Gravitar & 173.0 & 3351.4 \\
    Hero & 1026.97 & 30826.4 \\
    Ice Hockey & -11.15 & 0.9 \\
    James Bond & 29.0 & 302.8 \\
    Kangaroo & 52.0 & 3035.0 \\
    Krull & 1598.05 & 2665.5 \\
    Kung Fu Master & 258.5 & 22736.3 \\
    Montezuma Revenge & 0.0 & 4753.3 \\
    Ms Pacman & 307.3 & 6951.6 \\
    Name This Game & 2292.35 & 8049.0 \\
    Phoenix & 761.4 & 7242.6 \\
    Pitfall & -229.44 & 6463.7 \\
    Pong & -20.71 & 14.6 \\
    Private eye & 24.94 & 69571.3 \\
    Qbert & 163.88 & 13455.0 \\
    Riverraid & 1338.5 & 17118.0 \\
    Road Runner & 11.5 & 7845.0 \\
    Robotank & 2.16 & 11.9 \\
    Seaquest & 68.4 & 42054.7 \\
    Skiing & -17098.09 & -4336.9 \\
    Solaris & 1236.3 & 12326.7 \\
    Space Invaders & 148.3 & 1668.7 \\
    Star Gunner & 664.0 & 10250.0 \\
    Surround & -9.99 & 6.53 \\
    Tennis & -23.84 & -8.3 \\
    Time Pilot & 3568.0 & 5229.2 \\
    Tutankham & 11.43 & 167.6 \\
    Up n Down & 533.4 & 11693.2 \\
    Venture & 0.0 & 1187.5 \\
    Video Pinball & 0.0 & 17667.9 \\
    Wizard of Wor & 563.5 & 4756.5 \\
    Yars Revenge & 3092.91 & 54576.9 \\
    Zaxxon & 32.5 & 9173.3 \\
    \bottomrule
    \end{tabular}
        
\end{table}

\section{Log Residuals as an Approximation to Relative Error}
\label{app:residual}

Here we show that log residuals are approximations to the relative error. For a residual using a biased log transform given by 
\begin{align}
\delta = \log(\epsilon+\hat{y})-\log (\epsilon + y)
\end{align}
where $y$ is the true value, and $\hat{y}$ the prediction, with $y, \hat{y} \in \mathbb{R} \ge 0$ and $\epsilon \in \mathbb{R} > 0$ we have
\begin{align}
\delta &:=  \log(\epsilon+\hat{y})-\log(\epsilon+y) \\
    &=      \log(\frac{\epsilon+\hat{y}}{\epsilon+y}) \\
    &=      \log(1+(\frac{\epsilon+\hat{y}}{\epsilon+y}-1)) \\
    &\approx\frac{\epsilon+\hat{y}}{\epsilon+y}-1  \text{\quad (for $\frac{\epsilon+\hat{y}}{\epsilon+y}$ close to $1$)} \\
    &=      \frac{\hat{y}-y}{\epsilon+y},
\end{align}

which, for $\epsilon \ll y$, approximates the relative error. When using logarithms other than the natural logarithm, the log residual $\delta$ must first be multiplied by the natural log of the base used.

\end{document}